%% file: main.tex
\pdfoutput=1

\documentclass[11pt]{article}

\usepackage[final]{acl}

\usepackage{times}
\usepackage{latexsym}

\usepackage[T1]{fontenc}

\usepackage[utf8]{inputenc}

\usepackage{microtype}

\usepackage{inconsolata}

\usepackage{comment}
\usepackage{todonotes}
\usepackage{tabularx}
\usepackage{multirow}
\usepackage{graphicx}
\usepackage{booktabs}
\usepackage{arydshln}
\usepackage{pifont}
\usepackage{xcolor}
\usepackage{tcolorbox}
\usepackage{wrapfig}

\title{\modelname{}: Using Extractable Symbolic Programs to Evaluate Mathematical Reasoning}

\author{Xiaodong Yu\thanks{Equal Contribution. Work done when authors were PhD students at UPenn.}$^{1,3}$ \enspace\enspace\enspace Ben Zhou$^{\ast1,4}$ \enspace\enspace\enspace Hao Cheng$^2$ \enspace\enspace\enspace Dan Roth$^1$\\
$^1$University of Pennsylvania \enspace\enspace\enspace $^2$Microsoft Research\\$^3$AMD \enspace\enspace\enspace $^4$Arizona State University \\
\url{https://github.com/CogComp/reasoning-eval}}

\newcommand{\modelname}{\texttt{ReasonAgain}}

\input{latex/Definitions}

\begin{document}
\maketitle
\begin{abstract}
Existing math datasets evaluate the reasoning abilities of large language models (LLMs) by either using the final answer or the intermediate reasoning steps derived from static examples.
However, the former approach fails to surface model's uses of shortcuts and wrong reasoning while the later poses challenges in accommodating alternative solutions.
In this work, we seek to use symbolic programs as a means for automated evaluation if a model can consistently produce correct final answers across various inputs to the program. 
We begin by extracting programs for popular math datasets (GSM8K and MATH) using GPT4-o.
For those executable programs verified using the original input-output pairs, they are found to encapsulate the proper reasoning required to solve the original text questions.
We then prompt GPT4-o to generate new questions using alternative input-output pairs based the extracted program.
We apply the resulting datasets to evaluate a collection of LLMs.
In our experiments, we observe significant accuracy drops using our proposed evaluation compared with original static examples, suggesting the fragility of math reasoning in state-of-the-art LLMs.

\end{abstract}

\input{latex/sections/intro_v2}

\section{Methods}
\label{sec:method}

Assessing the reasoning capabilities of large language models (LLMs) presents significant challenges, primarily because the reasoning process is not consistently articulated, and standardizing its representation is difficult. Moreover, multiple reasoning paths may exist to arrive at the same solution. Consequently, it is impractical to simply output the reasoning process and evaluate its correctness directly. Typically, the accuracy of reasoning is assessed through question-answering formats, such as verifying the accuracy of an answer to a mathematical problem. However, this paper contends that relying solely on a single question-answer pair is inadequate for genuinely assessing reasoning capabilities because: 1) an incorrect reasoning path may coincidentally yield the correct answer, and 2) potential data contamination could enable models to memorize answers without engaging in a legitimate reasoning process. To effectively evaluate the reasoning abilities of LLMs, we introduce \modelname{}, which conceptualizes the reasoning process within Python code and automatically generates five additional perturbations of the same question. These perturbations retain the original reasoning process but feature different input values, thereby testing whether the model genuinely employs a correct reasoning process. The pipeline of \modelname{} is illustrated in Figure \ref{fig:pipeline}.

\paragraph{Encapsulating the reasoning process.} To explicitly represent the reasoning process of a math question, we first ask a pivot LLM (GPT-4o) to generate the parameters of questions. 

\begin{tcolorbox}[title=\footnotesize Generate Parameters of the Question,top=1mm,bottom=1mm]
\scriptsize
Identify numerical values in the given question, then replace some of them with Python parameters that are either int or float, so that the resulting abstract question is still answerable with the same general solution as the original question. Follow the the provided examples.\\

\{Few-shots examples\}\\

\{Question\}
\end{tcolorbox}

Then we use the generated parameter names to replace all the values in the question, and ask the LLM to generate a Python function that uses the generated parameters as the input to solve the question.

\begin{tcolorbox}[title=\footnotesize Generate Python Function of the Question,top=1mm,bottom=1mm]
\scriptsize
Write a Python program to solve the given abstract math question. Your program must contain a function called 'answer' that accepts the input parameters as specified in the question.\\

\{Few-shots examples\}\\

\{Question with parameters.\}

\end{tcolorbox}

After generating Python code for all the questions, to ensure the quality of the code, we first filter out all the code that cannot be compiled. Then we run the code by inputting the original parameter values, and we only keep the code that can output the correct answer.

\paragraph{Generate the perturbations of the question.} To generate the perturbations of the question, we first ask the model to generate 5 kinds of new parameter values given the original parameters using the following prompt. 

\begin{tcolorbox}[title=\footnotesize Prompt for Generating Alternative Parameter Values,top=1mm,bottom=1mm]
\scriptsize
Here is a math question with the parameter and parameter values. Please perturb the value of parameters into different values. Output five kinds of new values in the same format as the given parameters in five lines without index.\\

Question: \{Question\}\\

Parameters: \{Parameters\}
\end{tcolorbox}

Once we obtain these new parameter values, we prompt the model to update all values in the question to generate the corresponding new questions.

\begin{tcolorbox}[title=\footnotesize Prompt for Generating New Questions,top=1mm,bottom=1mm]
\scriptsize
Here is a math question with old parameter values, and five kinds of new parameter values. Please rewrite the question five times to update all the parameters from old value to each corresponding new value in five lines without index.\\

Question: \{Question\}\\

Old Parameters: \{Parameters\}\\

New Parameters: \{New Parameters\}\\

New Questions:

\end{tcolorbox}

To get the answers for each new question, we run the Python code for each set of new parameter values, and use the code's output as the target answer. To examine the robustness of models' reasoning capabilities, we then have the models answer the new questions and compare the outputs to the target answers.

\begin{table*}[t!]
\small
\setlength{\tabcolsep}{5pt}
\centering
\begin{tabular}{l|l|ccp{0.04\textwidth}p{0.04\textwidth}c|ccp{0.04\textwidth}p{0.04\textwidth}c}
\hline
\multicolumn{1}{c|}{\multirow{3}{*}{Model}} & \multirow{3}{*}{Prompt} & \multicolumn{5}{c|}{GSM8K}                                                                                                                            & \multicolumn{5}{c}{MATH}                                                                                                                             \\ \cline{3-12} 
\multicolumn{1}{c|}{}                       &                         & \multicolumn{2}{c|}{Accu.}                              & \multicolumn{3}{c|}{Normalized Accu.}                                                       & \multicolumn{2}{c|}{Accu.}                              & \multicolumn{3}{c}{Normalized Accu.}                                                       \\
\multicolumn{1}{c|}{}                       &                         & \multicolumn{1}{c}{Before} & \multicolumn{1}{c|}{After} & \multicolumn{1}{p{0.04\textwidth}}{Before} & \multicolumn{1}{p{0.04\textwidth}}{After} & \multicolumn{1}{c|}{\% of Correct} & \multicolumn{1}{c}{Before} & \multicolumn{1}{c|}{After} & \multicolumn{1}{p{0.04\textwidth}}{Before} & \multicolumn{1}{p{0.04\textwidth}}{After} & \multicolumn{1}{c}{\% of Correct} \\ \hline
\multirow{3}{*}{Llama3.1-8B}                & Direct                  & 21.59                      & \multicolumn{1}{l|}{7.05}  & 100                        & 34.27                     & 5.39                               & 20.88                      & \multicolumn{1}{l|}{14.30} & 100                        & 32.31                     & 9.62                              \\
                                            & CoT                     & 88.26                      & \multicolumn{1}{l|}{69.62} & 100                        & 75.31                     & 48.63                              & 71.49                      & \multicolumn{1}{l|}{44.02} & 100                        & 54.04                     & 33.71                             \\
                                            & CoT+SC                  & 85.75                      & \multicolumn{1}{l|}{68.01} & 100                        & 71.95                     & 39.92                              & 69.48                      & \multicolumn{1}{l|}{43.13} & 100                        & 53.06                     & 25.43                             \\ \hline
\multirow{3}{*}{Qwen2.5-7B}                 & Direct                  & 38.44                      & \multicolumn{1}{l|}{22.20} & 100                        & 42.78                     & 13.29                              & 35.34                      & \multicolumn{1}{l|}{19.52} & 100                        & 38.41                     & 13.64                             \\
                                            & CoT                     & 60.04                      & \multicolumn{1}{l|}{49.48} & 100                        & 63.79                     & 30.44                              & 38.96                      & \multicolumn{1}{l|}{25.30} & 100                        & 46.19                     & 17.53                             \\
                                            & CoT+SC                  & 68.39                      & \multicolumn{1}{l|}{56.09} & 100                        & 64.64                     & 30.01                             & 40.56                      & \multicolumn{1}{l|}{26.43} & 100                        & 44.95                     & 16.83                             \\ \hline
\multirow{3}{*}{GPT4o}                      & Direct                  & 66.57                      & \multicolumn{1}{l|}{52.93} & 100                        & 72.89                     & 48.86                              & 57.83                      & \multicolumn{1}{l|}{37.27} & 100                        & 56.25                     & 36.11                             \\
                                            & CoT                     & 93.73                      & \multicolumn{1}{l|}{\textbf{75.68}} & 100                        & \textbf{79.52}                     & \textbf{58.80}                              & 84.34                      & \multicolumn{1}{l|}{\textbf{50.76}} & 100                        & 55.62                     & 34.29                             \\
                                            & CoT+SC                  & 94.44                      & \multicolumn{1}{l|}{74.87} & 100                        & 78.05                     & 55.12                              & 82.33                      & \multicolumn{1}{l|}{50.36} & 100                        & 55.41                     & 27.80                             \\ \hline
\multirow{3}{*}{GPT4-Turbo}                 & Direct                  & 45.43                      & \multicolumn{1}{l|}{35.04} & 100                        & 56.13                     & 26.63                              & 47.39                      & \multicolumn{1}{l|}{31.16} & 100                        & 45.76                     & 22.88                             \\
                                            & CoT                     & 54.75                      & \multicolumn{1}{l|}{43.49} & 100                        & 70.02                     & 48.12                              & 36.14                      & \multicolumn{1}{l|}{28.19} & 100                        & \textbf{61.78}                     & \textbf{42.22}                             \\
                                            & CoT+SC                  & 51.52                      & \multicolumn{1}{l|}{40.95} & 100                        & 71.23                     & 50.09                              & 55.02                      & \multicolumn{1}{l|}{37.59} & 100                        & 55.62                     & 32.85                             \\ \hline
\end{tabular}
\caption{Performance of LLMs on GSM8K, MATH and corresponding perturbations generated by \modelname. "Normalized Accu." refers to the performance on the subset of the test cases that the model answers correctly before perturbation. "Before" refers to the performance on the original dataset. "After" refers to the performance on the perturbations. "\% of Correct" refers to the percentage of the cases that the model solves all the perturbations correctly. The final metric reflects whether the evaluated LLMs truly understand the necessary reasoning.}
\label{tab:main_result}
\end{table*}

\section{Experiments}
\label{sec:exp}

\subsection{Experiment Settings}
\paragraph{Datasets.} We sample 2k questions from GSM8k \citep{cobbe2021gsm8k} and 1k questions from MATH \citep{hendrycksmath2021}. As discussed in Section \ref{sec:method}, we first ask the model to generate the Python code for each question, and then we filter out all the problematic code that cannot be compiled or fail to return the correct gold answer. After filtering, in total, we have 1121 cases from GSM8k, and 268 cases from MATH. For each case, we use \modelname~to generate 5 perturbations as the new test cases, which gives us 5605 cases for GSM8k, and 1072 cases for MATH. We use GPT-4o \citep{openai2024gpt4technicalreport} as the pivot LLM to generate all the parameters, code, and perturbations.

\paragraph{Baselines.} We evaluate 4 LLMs in this paper: GPT-4-Turbo \citep{openai2024gpt4technicalreport}, GPT-4o \citep{openai2024gpt4technicalreport}, LLama-3.1-8B \citep{dubey2024llama3herdmodels}, and Qwen-2.5-7B \cite{qwen2.5} using the following different prompting settings: direct, few-shot Chain-of-thought (CoT) \cite{wei2022chain}, and few-shot Chain-of-thought + self-consistency (CoT+SC) \cite{wang2022self}.

\noindent\textbf{Direct:} We ask the model to directly answer the question without providing any examples using the following prompts.

\begin{tcolorbox}[title=\footnotesize Prompt for Generating Alternative Parameter Values,top=1mm,bottom=1mm]
\scriptsize
Answer the math question below. Only output the answer without units and any context words.\\

Question: \{Question\}\\

Answer:
\end{tcolorbox}

\noindent\textbf{Few-shot CoT:} We follow the same CoT template and the same 8-shot math examples from \citet{wei2022chain}. Temperature is set to 0. 

\noindent\textbf{Few-shot CoT+SC:} Following \citet{wang2022self}, temperature is set to 0.7, and we run each query 5 times. The majority of the outputs will be used as the final answer.  

\paragraph{Evaluation Metrics.} We report Exact Match accuracy (EM) for all the experiments. Predicted answers are parsed by CoT format, and we round both gold answers and predicted answers before checking if the values are same.

\subsection{Main Results}
\label{ssec:results}

We show our main experiment results using our proposed \modelname{} evaluation pipeline in \autoref{tab:main_result}. We observe a substantial performance drop across all models on both GSM8K and MATH. For direct inference, models experience 10\%-15\% drop in performance, regardless of their size and capabilities.
The decline is not mitigated by chain-of-thought and self-consistency inference methods, as we observe a similar 10\% to 20\% drop after our perturbation. 
In the normalized accuracy results, we show that models often demonstrate a misleading impression of their performances: they only answer 50\% to 80\% of the perturbed questions correctly on the questions that they initially answered correctly.
A more concerning finding is that models only truly understand at most half of the questions, and sometimes even less than 30\%, as suggested by the ``\% of Correct'' results.
Combining these findings, we contend that \modelname{} is an effective method for evaluating the true capabilities of models in mathematical reasoning, revealing that current models' performances are overestimated by previous evaluation methods solely based on static data.

\subsection{Human Evaluation}
\label{sec:human}
To assess whether the generated code accurately embodies a valid reasoning process, we randomly sample 200 cases from GSM8K and MATH (100 each), and ask three human experts to judge the correctness of our generated perturbations.
Specifically, the annotators are asked to understand the generated code, and check the correctness of the target answers of perturbations.
In summary, we find 8 of the 100 cases from GSM8K and 17 of the 100 cases from MATH contain errors.
These issues are mainly due to some positive parameters being negative or the model failing to generate the correct program that encapsulates the necessary reasoning process, which can be potential directions for further improvements.
Despite these errors, the majority of our new test cases remain valid and useful for proper evaluation purposes.

\section{Related Work}
Many works have discussed language model bias and inconsistency during reasoning~\cite{li2024deceptive, li2024famicom, zhou2024conceptual} and adversarial and contrastive evaluation~\cite{gardner2020evaluating, patel2021nlp, yu-etal-2024-reeval}.
Here, we provide a novel way for automatic mathematical reasoning evaluation by checking the reasoning reliability using alternative input-output pairs with the same text question context.
While previous studies have successfully used decomposed methods to solve math questions more reliably~\cite{hao2023reasoning, madaanself, gao2023pal, xia2024evaluating}, our work highlights the reasoning challenges faced by existing LLMs. 
This indicates a need for more advanced developments to further improve the reliability of LLMs in mathematical reasoning. 
Another related line of work~\citep[][\textit{inter alia}]{xia2024evaluating} aims to surface the reasoning flaws of LLMs by examining their intermediate steps (\eg CoT processes).
In contrast, we bypasses the process evaluation and instead evaluate whether the model truly understand how to solve a problem by checking the consistency of its answers using the same reasoning process encapsulated in a symbolic program.
We have noticed a contemporary work \cite{mirzadeh2024gsm} that also generates perturbations of math questions to evaluate the LLMs' mathmatical reasoning capabilities.
However, while \citet{mirzadeh2024gsm} uses symbolic templates to create perturbations, we leverage Python code extracted by LLMs in an automatical fashion. 

\section{Conclusion}
In this work, we propose \modelname{}, a novel evaluation method to better benchmark large language models' true capabilities on mathematical reasoning.
\modelname{} employs a symbolic program-based perturbation method that changes the numerical values in the original math questions and derives the corresponding target answers. We then evaluate models on such perturbed questions. Experiments show that representative SoTA LLMs perform significantly worse on our modified questions, suggesting that 1) existing models do not truly understand the reasoning process behind math questions, even when they occasionally predict the correct answer; 2) existing static data based evaluation methods are inadequate, leading to an overly optimistic perception of model performances in mathematical reasoning. \modelname{} offers a more effective alternative for evaluating LLMs' reasoning capabilities.

\section*{Limitations}
Our work has several limitations. 

\paragraph{Imperfect Programs.} As pointed out in \S\ref{sec:human}, some mistakes exist in the current generated programs, which leads to partially incorrect gold labels in some perturbed questions. We will explore better filtering mechanisms in later versions. However, such mistakes do not impact our overall conclusion, as model performances are much lower than the upper bounds. 

\paragraph{Limited Program Coverage.} Our program generation is limited by a conceptualization process proposed in \citet{zhou2024conceptual}, which does not work well on certain types of math questions, such as geometry-related ones. As a result, \modelname{} only works on a subset of all existing math questions.

\paragraph{Limited Reasoning Types.} Our general formulation can be applied to other reasoning types, such as multiple-choice questions. However, we only focus on math questions in this work.

\bibliography{latex/custom}




\end{document}

%% file: latex/sections/intro_v2.tex
\section{Introduction}

Mathematical reasoning is a fundamental skill essential for numerous complex applications, leading to a recent growing research effort on advancing large language models (LLMs) in this area.
Thus, proper evaluation of LLMs' mathematical reasoning is crucial.
Most previous studies have primarily evaluated LLMs using static datasets, such as GSM8K \citep{cobbe2021gsm8k} and MATH \citep{hendrycksmath2021}.
Typically, evaluations focus solely on the final answers, overlooking reasoning flaws~\citep{lewkowycz2022solvingquantitativereasoningproblems} and potential data contamination issues. 
Despite impressive results, LLMs can reply on shortcuts rather than true reasoning, displaying high sensitivity to input tokens~\citep{li2024deceptive,li2024famicom}.
Alternatively, some works \citep{sawada2023arbadvancedreasoningbenchmark,golovneva2023roscoesuitemetricsscoring} use model-based techniques to assess the reasoning quality, but these can suffer from model biases, limiting accommodation for alternative solutions.

In this paper, we present a focused study on evaluating mathematical reasoning which can be concisely encapsulated by symbolic programs, \ie Python programs.
For such cases, we can automatically generate a diverse set of new test cases (input-output pairs) by varying the valid inputs fed into the program.
Thus, if LLMs truly employ the appropriate reasoning process (as embodied by the programs) to solve the original question, they should also be able to consistently solve all new test cases.
This approach allows us to evaluate the reasoning quality directly by examining the final answers, without ruling out alternatives.

To avoid costly manual annotations, we use the state-of-the-art (SoTA) LLM (GPT4-o) to generate Python programs for GSM8K and MATH.
We retain only those questions with \textit{extractable} programs, which can be automatically validated for our evaluation.
This means the programs can be executed to produce the original gold answers.
Upon manual inspection, $92\%$ and $83\%$ of the programs from GSM8K and MATH genuinely demonstrate the correct reasoning process required to solve the original questions.
We then prompt GPT4-o to propose alternative valid inputs based on the extracted program and the original question.
These inputs are then used to generate new input-output pairs derived from the program.
Finally, GPT4-o is tasked to update the original question using these proposed inputs to create new test cases for evaluation.

\begin{figure*}[t!]
    \centering
    \includegraphics[width=\textwidth]{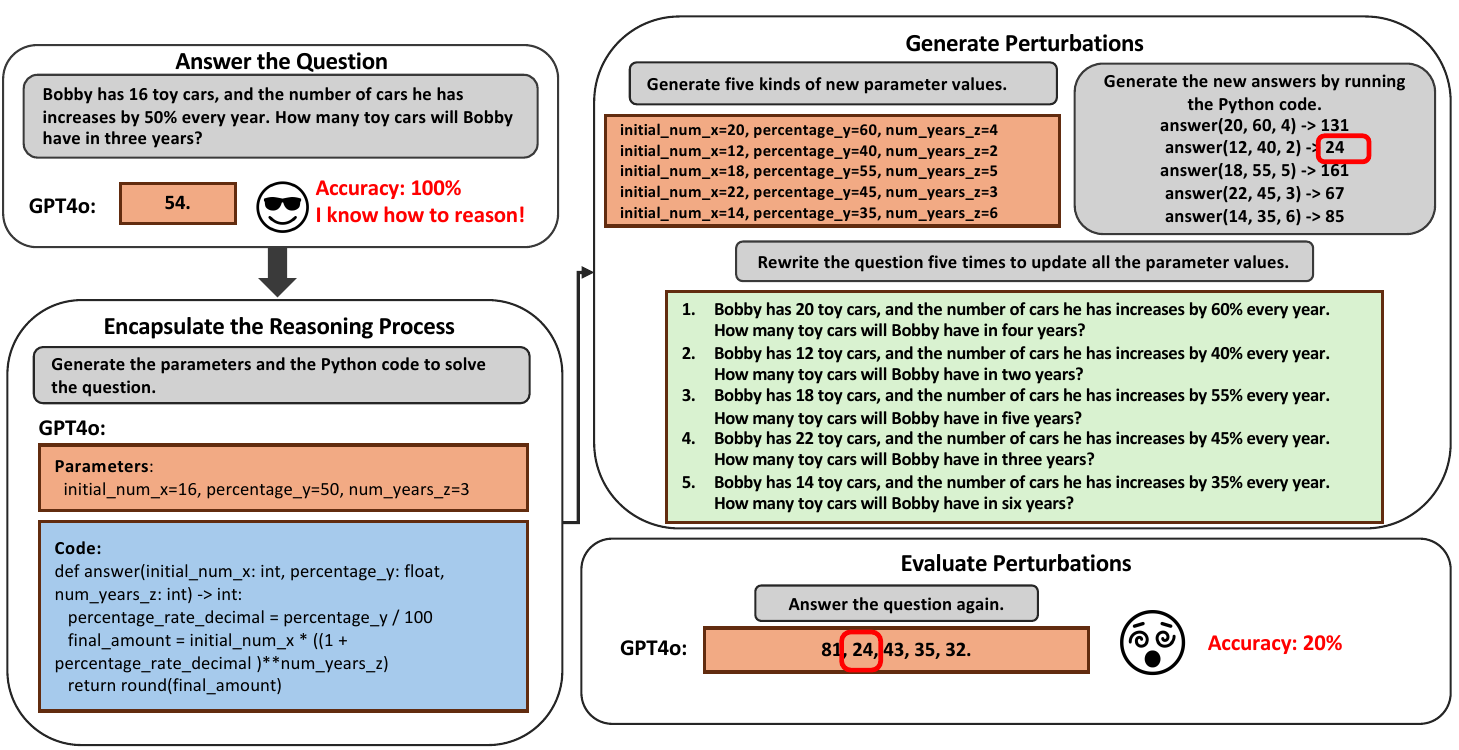}
    \caption{Pipeline of \modelname, and a running example perturbed by \modelname. }
    
    \label{fig:pipeline}
\end{figure*}

Our experiments reveal significant declines in the performance of SoTA LLMs when evaluated on our generated data. For example, for questions that GPT-4-turbo can answer correctly, over half of the alternatives generated by our method can not be properly solved. This highlights the fragility of the mathematical reasoning capabilities of existing LLMs. In contrast to traditional static data evaluation methods, our proposed approach, \modelname{}, offers a viable solution for identifying these weaknesses and providing a reliable evaluation of reasoning abilities.